\documentclass{article} 
\usepackage{iclr2016_conference,times}

\usepackage{algorithm}
\usepackage{algorithmic}
\usepackage{color}
\usepackage{amsmath}
\usepackage{amsfonts}
\usepackage{graphicx} 
\usepackage{epstopdf}
\usepackage{caption}
\usepackage{subcaption} 
\usepackage{tikz}
\usetikzlibrary{arrows}
\usepackage{verbatim}
\usepackage{hhline}
\usepackage{setspace}
\usepackage{url}
\usetikzlibrary{arrows}

\tikzstyle{box}=[draw, thick,minimum size=2em]
\tikzstyle{cir}=[circle,thick,draw,minimum size=2em]
\tikzstyle{init} = [pin edge={to-,thin,black}]
\title{MuProp: Unbiased Backpropagation for Stochastic Neural Networks}

\author{
Shixiang Gu$^1$ $^2$, Sergey Levine$^3$, Ilya Sutskever$^3$, and Andriy Mnih$^4$\\
$^1$ University of Cambridge\\
$^2$ MPI for Intelligent Systems, T\"{u}bingen, Germany \\
$^3$ Google Brain\\
$^4$ Google DeepMind \\
\texttt{\{shanegu,slevine,ilyasu,amnih\}@google.com}
}

\iclrfinalcopy 

\begin{document}

\maketitle

\begin{abstract}
Deep neural networks are powerful parametric models that can be trained efficiently using the backpropagation algorithm. Stochastic neural networks combine the power of large parametric functions with that of graphical models, which makes it possible to learn very complex distributions. However, as backpropagation is not directly applicable to stochastic networks that include discrete sampling operations within their computational graph, training such networks remains difficult. We present MuProp, an unbiased gradient estimator for stochastic networks, designed to make this task easier. MuProp improves on the likelihood-ratio estimator by reducing its variance using a control variate based on the first-order Taylor expansion of a mean-field network. Crucially, unlike prior attempts at using backpropagation for training stochastic networks, the resulting estimator is unbiased and well behaved. Our experiments on structured output prediction and discrete latent variable modeling demonstrate that MuProp yields consistently good performance across a range of difficult tasks.
\end{abstract}

\section{Introduction}

Deep neural networks \citep{krizhevsky2012imagenet,hinton2012deep,sutskever2014sequence} are responsible for numerous state-of-the-art results in a variety of domains, including computer vision, speech recognition, and natural language processing.
The cornerstone of their success has been the simple and scalable backpropagation algorithm \citep{rumelhart1986learning}.
Backpropagation provides an efficient way of computing the derivatives of the error with respect to the model parameters.
The key to the applicability of backpropagation is to utilize deterministic, differentiable activation functions, such as sigmoidal and softmax units, to make the model differentiable.
However, in certain real-world scenarios, it is more suitable to learn a model that can carry out a sequence of stochastic operations internally, in order to represent a complex stochastic process. Such stochastic networks are studied in policy gradient reinforcement learning methods \citep{williams1992simple,weaver2001optimal,peters2006policy}, probabilistic latent variable models for structured prediction, unsupervised learning of generative models~\citep{tang2013learning,kingma2013auto}, and most recently, attention and memory networks~\citep{mnih2014recurrent,zaremba2015reinforcement}.



The versatility of stochastic neural networks motivates research into more effective algorithms for training them. Models with continuous latent variables and simple approximate posteriors can already be trained efficiently using the variational lower bound along with the reparameterization trick, which makes it possible to train both the model and the inference network using backpropagation \citep{kingma2013auto,rezende2014stochastic}. Training models with discrete latent variable distributions, such as Bernoulli or multinomial, is considerably more difficult. Unbiased estimators based on the likelihood-ratio method tend to be significantly less effective than biased estimators, such as the straight-through method ~\citep{bengio2013estimating,raiko2014techniques} and the estimator proposed by ~\cite{gregor2014deep}. We hypothesize that this is due to the fact that, unlike the biased estimators, the unbiased ones do not take advantage of the gradient information provided by the backpropagation algorithm. However, the biased estimators are heuristic and not well understood, which means that it is difficult to enumerate the situations in which these estimators will work well. We posit that an effective method for training stochastic neural networks should take advantage of the highly efficient backpropagation algorithm, while still providing the convergence guarantees of an unbiased estimator.


To that end, we derive MuProp, an unbiased gradient estimator for deep stochastic neural networks that is based on backpropagation. To the best of our knowledge, it is the first unbiased estimator that can handle both continuous and discrete stochastic variables while taking advantage of analytic gradient information. MuProp's simple and general formulation allows a straightforward derivation of unbiased gradient estimators for arbitrary \textit{stochastic computational graphs} -- directed acyclic graph (DAG) with a mix of stochastic and deterministic computational nodes (see, e.g., \cite{schulman2015gradient}). While the algorithm is applicable to both continuous and discrete distributions, we used only discrete models in our experiments, since the reparameterization trick~\citep{kingma2013auto} already provides an effective method for handling continuous variables. We present experimental results for training neural networks with discrete Bernoulli and multinomial variables for both supervised and unsupervised learning tasks. With these models, which are notoriously difficult to train, biased methods often significantly outperform the unbiased ones~\citep{dayan1995helmholtz,gregor2014deep, raiko2014techniques}, except in certain cases~\citep{mnih2014neural}. Our results indicate that MuProp's performance is more consistent and often superior to that of the competing estimators. It is the first time that a well-grounded, unbiased estimator consistently performs as well or better than the biased gradient estimators across a range of difficult tasks.

\section{Related Work}
\label{sec:bg}

Probabilistic latent variable models described by neural networks date back to pioneering early work on sigmoidal belief networks and Helmholtz machines \citep{neal1992connectionist,dayan1995helmholtz}. However, their adoption has been hindered by the lack of training methods that are both efficient and theoretically sound. Algorithms based on Markov chain Monte Carlo \citep{neal1992connectionist} and mean-field inference \citep{saul1996mean} are theoretically well-grounded, but do not scale well to large models. The wake-sleep algorithm \citep{hinton1995wake} scales to large models, but does not optimize a well defined-objective function, and therefore does not provide convergence guarantees. Recently, several new scalable methods have been proposed for training such models.
A number of these algorithms use likelihood ratio estimators with variance reduction techniques \citep{ranganath2014black,mnih2014neural, gregor2014deep}, inspired by methods from reinforcement learning \citep{williams1992simple,weaver2001optimal,peters2006policy}.
Another group of algorithms specifically addresses structured predictions tasks involving binary latent variables. Instead of performing inference using an inference network, these algorithms either use importance sampling to reweight samples from the prior~\citep{tang2013learning,raiko2014techniques}, or rely on heuristics for approximate backpropagation through the stochastic units \citep{bengio2013estimating,raiko2014techniques,gregor2014deep}.  

\subsection{Likelihood-Ratio Gradient Estimation}
\label{sec:lr}
Consider a simple stochastic system with discrete random variable $x$ whose probability is given by $p_\theta(x)$, and a loss function $f(x)$. This formulation subsumes the training objectives of popular generative models such as sigmoid belief networks, with or without inference networks. The objective of training is to minimize the expected cost $L(\theta) = \mathbb{E}_{p_\theta(x)}[f(x)]$. The gradient $g=\partial L / \partial \theta$ is usually intractable to compute exactly, and must therefore be estimated using an estimator $\hat{g}(x)$, i.e. $g \approx \sum_{i=1}^{m} \hat{g}(x_i) / m$, where $x_i\sim p_\theta(x)$ and $m$ is the number of Monte Carlo samples.

The likelihood-ratio (LR) estimator, of which the popular REINFORCE algorithm is a special case~\citep{williams1992simple,peters2006policy}, provides a convenient method for estimating this gradient, and serves as the basis for all of the unbiased estimators discussed in this paper, including MuProp.
The LR method only requires that $p_\theta(x)$ is differentiable with respect to $\theta$:        
\begin{equation} \label{eq:deriv_est_reinforce}
 \begin{split}
 g_{(\mathcal{LR})} &= \nabla_{\theta} \mathbb{E}_{p_\theta(x)}[f(x)]=  \sum_x \nabla_{\theta} p_\theta(x) f(x)\\
 &=  \sum_x p_\theta(x) \nabla_{\theta} \log p_\theta(x)\cdot f(x)= \mathbb{E}_{p_\theta(x)}[\nabla_{\theta} \log p_\theta \cdot f(x)]\\
  \hat{g}_{(\mathcal{LR})} 
  &= \nabla_{\theta}\log p_{\theta}(x)\cdot f(x) \quad \textrm{where } x\sim p_\theta\\
  \end{split}
\end{equation}
In its basic form, this estimator is unbiased, but tends to have high variance, and variance reduction techniques are typically required to make it practical~\citep{williams1992simple,peters2006policy}. A major deficiency of the LR estimator is that it fails to utilize information about the derivative of the cost function, unlike the most successful biased estimators. We will show that MuProp, which combines the derivative information with the LR estimator, can outperform these biased estimators, as well as the standard LR method.


\subsection{Variance Reduction with the Likelihood-Ratio Estimator}
\label{sec:baseline}

High variance of LR estimators can make convergence very slow or impossible. This is further exacerbated by the fact that LR estimators for minibatch training of neural networks typically use $m=1$, in order to maximize the variety of examples seen by the network~\citep{kingma2013auto,gregor2014deep,rezende2014stochastic,mnih2014neural}. The simplest way to reduce variance is to increase $m$, but this is computationally expensive. Efficient and effective variance reduction techniques are therefore crucial for the LR estimator to be practical.


The derivation of MuProp in Section~\ref{sec:muprop} uses a variance reduction technique known as control variates \citep{paisley2012variational}.
The main idea is to subtract an analytically tractable term from the LR estimate in order to reduce the variance of the Monte Carlo estimate, and then add back the analytical expectation of this term to recover an unbiased estimator:
\begin{equation} \label{eq:baseline}
 \begin{split}
  \mathbb{E}_{p_{\theta}(x)}[\nabla_{\theta}\log p_{\theta}(x)\cdot f(x)] &= \mathbb{E}_{p_{\theta}(x)}[\nabla_{\theta}\log p_\theta(x)\cdot (f(x)-b-h(x))] + \mu   
 \end{split}
\end{equation}
In this example, $b + h(x)$ is a control variate, which is also known as a sample-dependent baseline. The expectation of the baseline needs to be added to the expression to make the resulting estimator unbiased. If the baseline is constant (i.e. $h(x)=0$), its contribution to the gradient estimate is always zero in expectation, since $\sum_x p_\theta(x) \nabla_{\theta} \log p_\theta(x)  = \sum_x \nabla_{\theta} p_\theta(x) = \nabla_{\theta} \sum_x p_\theta(x) = \nabla_{\theta} 1 = 0$. Otherwise, we must compute the expectation of the baseline $(b+h(x))\nabla \log p_\theta(x)$ analytically, as $\mu = \mathbb{E}_{p_\theta(x)}[\nabla_{\theta}\log p_\theta(x)\cdot h(x)]$.  
In our experiments, we utilize three types of variance reduction techniques for the estimators that include the same term as Eq.~\ref{eq:baseline}. While these techniques, proposed in a similar context in \citep{mnih2014neural}, do not result in the optimal variance reduction, they work well enough to make LR estimators useful in practice.\footnote{Optimal baselines are somewhat involved, as they take into account the magnitude of $\nabla_{\theta}\log p_\theta(x)$ and are different for each parameter \citep{weaver2001optimal}.} For the precise details of these variance reduction techniques, please refer to~\cite{mnih2014neural}.  For reference, the techniques are: 

\begin{itemize}
  \item \textbf{Centering the learning signal (C)}, which involves subtracting from the learning signal its moving average (corresponding to $b$ in Eq.~\ref{eq:baseline}).
 \item \textbf{Input-dependent baseline (IDB)}, which allows baseline $b$ to depend on the input or the sample from the previous layer ($x_0$). This baseline is a neural network parameterized by $\psi$, and is trained jointly with the model to minimize the expected square of the centered learning signal $\mathbb{E}_{p_\theta(x|x_0)}[(l(x)-b-B_\psi(x_0))^2]$.
 \item \textbf{Variance normalization (VN)}, which keeps track of the moving average of the signal variance $v$, and divides the learning signal by $\max(1,\sqrt{v})$. Unlike the other two techniques, this does not correspond to a baseline, and is a type of adaptive gradient.
\end{itemize}

\section{MuProp}
\label{sec:muprop}
The MuProp estimator has two components: a deterministic term $\hat{g}_{MF}$, which is computed by backpropagation through a \emph{mean-field network} that we describe in Section~\ref{sec:mfnet}, and a LR term $\hat{g}_{R}$, which accounts for the residuals to produce an unbiased gradient estimate. In the following, let $\mu_x(\theta)$ be $\mathbb{E}_{p_\theta(x)}[x]$ and $\bar{x}$ be any value that does not depend on $x$.  The key idea of MuProp is summarized in the following equation:
\begin{equation} \label{eq:deriv_est}
 \begin{split}
  \hat{g}_{(\mathcal{\mu})} 
  &= \underbrace{\nabla_{\theta} \log p_\theta(x) \cdot [f(x) - f(\bar{x}) - f'(\bar{x})(x-\bar{x})]}_\textrm{$\hat{g}_{R}$} + 
  \underbrace{f'(\bar{x})\nabla_{\theta} \mu_x(\theta)}_\textrm{$\hat{g}_{MF}$} 
  \end{split}
\end{equation}
\noindent where $x\sim p_\theta$, and $\hat{g}_{(\mathcal{\mu})}$ is an unbiased estimate of the gradient. The derivation follows directly from the baseline technique described in Section~\ref{sec:baseline}: MuProp uses a control variate that corresponds to the first-order Taylor expansion of $f$ around some fixed value $\bar{x}$, i.e., $h(x) = f(\bar{x}) + f'(\bar{x})(x-\bar{x})$. The idea of using Taylor expansion as a baseline is also explored by~\citep{paisley2012variational} and \citep{gregor2014deep}; however, \citep{paisley2012variational} does not explore the estimator in the context of arbitrary stochastic computational graphs, and \citep{gregor2014deep} chooses a different form of Taylor expansion that makes the estimator biased. The latter approach also does not generalize to multi-layer graphs.

Our contribution with MuProp is to extend the idea of Taylor expansions as baselines and make them applicable to arbitrary stochastic computational graphs. We will describe our derivation and design choices, as well as possible extensions to the standard form of MuProp that can enhance the statistical efficiency of the method. In the appendix, we also present a simple algorithm to compute the MuProp estimator using automatic differentiation, which makes it easy to integrate into any existing automatic differentiation library.

\subsection{Generalized Backpropagation with Mean-Field Networks}
\label{sec:mfnet}

\begin{figure}
\centering
\begin{subfigure}{.39\textwidth}
\centering
\resizebox{0.5\textwidth}{0.75\textwidth}{
\begin{tikzpicture}[auto,>=latex']
  \node [cir] (c) {$s$};
    \node [cir] (a) [below of=c, node distance=1.5cm]{$x$};
    \node [cir] (b) [below of=a, node distance=1.5cm]{$h$};
    \node (start) [below of=b, node distance=1.0cm, coordinate]{};
        \node  (end) [above of=c,node distance=1.5cm] {$f(s)$};
  \path[->, blue] (c) edge node {} (end);
    \path[->, blue] (a) edge node {} (c);
    \path[->, blue] (b) edge node {$\theta$} (a);
    \path[->, blue] (start) edge node {} (b);
    
     \node [cir] (c2) [right of=c, node distance=3 cm]{$\bar{s}$};
      \node [cir] (a2) [below of=c2, node distance=1.5cm]{$\bar{x}$};
    \node [cir] (b2) [below of=a2, node distance=1.5cm]{$\bar{h}$};
    \node (end2) [above of=c2, node distance=1.5cm]{$\bar{f}$};
    \node (start2) [below of=b2,node distance=1.0cm, coordinate] {};
  \path[->] (c2) edge node {} (end2);
    \path[->] (a2) edge node {} (c2);
    \path[->] (b2) edge node {$\theta$} (a2);
    \path[->] (start2) edge node {} (b2);
    
    \path[->,transform canvas={xshift=-2.0pt}, red] (end2) edge node {} (c2);
    \path[->,transform canvas={xshift=-2.0pt}, red] (c2) edge node {} (a2);
    \path[->, red] (a2) edge node {$\frac{\partial \bar{f}}{\partial \bar{x}}$} (a);
    \path[->, red] (b2) edge node {$\frac{\partial \bar{f}}{\partial \bar{h}}$} (b);
    \path[->, red] (c2) edge node {$\frac{\partial \bar{f}}{\partial \bar{s}}$} (c);
    \path[->, transform canvas={xshift=-2.0pt}, red] (a) edge node {} (b);
       \path[->, transform canvas={xshift=-2.0pt}, red] (a2) edge node {} (b2);
          \path[->, transform canvas={xshift=-2.0pt}, red] (b2) edge node {} (start2);
            \path[->, transform canvas={xshift=-2.0pt}, red] (b) edge node {} (start);
           \path[->, transform canvas={xshift=-2.0pt}, red] (c) edge node {} (a);
  \path[->, transform canvas={xshift=-2.0pt}, red] (end) edge node {} (c);
\end{tikzpicture}
}
\end{subfigure}
\begin{subfigure}{.6\textwidth}
\centering
\resizebox{0.6\textwidth}{0.5\textwidth}{
\begin{tikzpicture}[auto,>=latex']

  \node [cir] (c) {$s$};
    \node [cir] (a) [below of=c, node distance=1.5cm]{$x$};
    \node [cir] (b) [below of=a, node distance=1.5cm]{$h$};
    \node (start) [below of=b, node distance=1.0cm, coordinate]{};
        \node  (end) [above of=c,node distance=1.5cm] {$f(s)$};
  \path[->, blue] (c) edge node {} (end);
    \path[->, blue] (a) edge node {} (c);
    \path[->, blue] (b) edge node {$\theta$} (a);
    \path[->, blue] (start) edge node {} (b);
    
    \node [cir] (c3) [right of=c, node distance=2cm]{$\bar{s}(x)$};
    \node (end3) [above of=c3, node distance=1.5cm]{$\bar{f}(x)$};
    \path[->, blue] (a) edge node {} (c3);
    \path[->] (c3) edge node {} (end3);
    
     \node [cir] (c2) [right of=c, node distance=4 cm]{$\bar{s}$};
      \node [cir] (a2) [below of=c2, node distance=1.5cm]{$\bar{x}(h)$};
    \node (end2) [above of=c2, node distance=1.5cm]{$\bar{f}(h)$};
  \path[->] (c2) edge node {} (end2);
    \path[->] (a2) edge node {} (c2);
    \path[->,blue] (b) edge node {} (a2);
    
     \node [cir] (c4) [right of=c, node distance=6 cm]{$\bar{s}$};
      \node [cir] (a4) [below of=c4, node distance=1.5cm]{$\bar{x}$};
    \node [cir] (b4) [below of=a4, node distance=1.5cm]{$\bar{h}$};
    \node (end4) [above of=c4, node distance=1.5cm]{$\bar{f}$};
    \node (start4) [below of=b4,node distance=1.0cm, coordinate] {};
  \path[->] (c4) edge node {} (end4);
    \path[->] (a4) edge node {} (c4);
    \path[->] (b4) edge node {} (a4);
    \path[->] (start4) edge node {} (b4);

    \path[->,transform canvas={xshift=-2.0pt}, red] (end2) edge node {} (c2);
    \path[->,transform canvas={xshift=-2.0pt}, red] (c2) edge node {} (a2);
    \path[->, red] (a2) edge node {$\frac{\partial \bar{f}}{\partial \bar{x}}$} (a);
    \path[->, red] (c3) edge node {$\frac{\partial \bar{f}}{\partial \bar{s}}$} (c);
    \path[->, red] (b4) edge node {$\frac{\partial \bar{f}}{\partial \bar{h}}$} (b);
       \path[->, transform canvas={xshift=-2.0pt}, red] (c4) edge node {} (a4);
     \path[->, transform canvas={xshift=-2.0pt}, red] (end4) edge node {} (c4);
     \path[->, transform canvas={xshift=-2.0pt}, red] (a4) edge node {} (b4);
     \path[->, transform canvas={xshift=-2.0pt}, red] (b4) edge node {} (start4);
            \path[->,transform canvas={xshift=-2.0pt}, red] (end3) edge node {} (c3);
    \path[->, transform canvas={xshift=-2.0pt}, red] (c) edge node {} (a);
     \path[->, transform canvas={xshift=-2.0pt}, red] (end) edge node {} (c);
     \path[->, transform canvas={xshift=-2.0pt}, red] (a) edge node {} (b);
     \path[->, transform canvas={xshift=-2.0pt}, red] (b) edge node {} (start);

\end{tikzpicture}
}
\end{subfigure}
\caption{Standard MuProp (left) and MuProp with \textit{rollout} (right). Circles indicate stochastic nodes. Blue, black, and red arrows indicate stochastic forward pass, deterministic mean-field forward pass, and gradient flow in backward pass respectively. $\bar{f}$ indicates mean-field evaluation of $f$.}
\label{fig:scg2}
\end{figure}
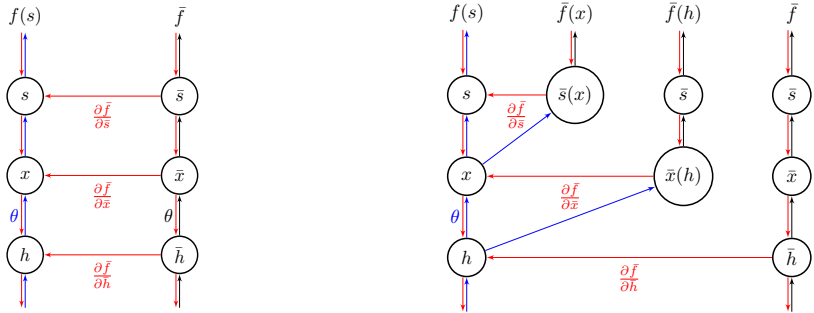

It is not always possible to directly apply Eq.~\ref{eq:deriv_est} to an arbitrary computation graph because of the $f'(\bar{x})$ terms. If the computation graph includes a discrete sampling operation as in Figure~\ref{fig:scg2}, then one cannot directly define continuous gradient through such an operation. As a solution, we use a deterministic mean-field network, which ignores the sampling operations in the original graph and propagates the mean values instead of the samples. Figure~\ref{fig:scg2} shows how the backward pass computes the gradient with respect to $\theta$. MuProp computes the Taylor expansion using the mean-field network, which is fully differentiable, and uses these terms to reduce the variance. While the choice to use the mean-field network seems arbitrary, we show that a proper recursive derivation of Eq.~\ref{eq:deriv_est} for deep networks naturally leads to recursive Taylor expansions around the mean functions. Furthermore, if $\bar{x}$ coincides with the mean-field forward pass, the gradient estimator simplifies to Eq.~\ref{eq:deriv_est}. The full derivation is given in the appendix.

\section{Comparison with Other Gradient Estimators}
\label{sec:estimators}

While MuProp is derived from the LR estimator, which is also unbiased, it is closely related to two biased estimators commonly used to train stochastic binary networks. In this section we will describe these two estimators, the Straight-Through estimator and the 1/2 estimator.  
To simplify notation, we describe the estimators assuming one stochastic variable $x$. The provided estimator expressions however generalize easily to arbitrary stochastic computational graphs.\footnote{Only the LR and MuProp estimators have a principled extension to deep networks. The other methods can be extended to deep networks heuristically.} The estimators mentioned above are summarized in Table~\ref{tab:grads} for completeness.

We do not consider the importance-sampling based approaches \citep{raiko2014techniques,tang2013learning,bornschein2014reweighted,burda2015importance} as these can be interpreted as optimizing a different objective \citep{burda2015importance} that requires sampling the latent variables multiple times. The resulting estimators are more specialized than the ones considered in this paper, which are agnostic to the objective being optimized. As a result, MuProp and the other general estimators can be applied to the objective optimized by the importance-sampling based methods. We leave exploring this direction as future work.

\subsection{The Straight-Through Estimator}
\label{sec:st}
The straight-through estimator (ST)~\citep{bengio2013estimating,raiko2014techniques} is a biased but low-variance estimator, devised primarily for binary stochastic neurons\footnote{The ST estimator was proposed by~\cite{hinton2012coursera} and named in~\cite{bengio2013estimating}. The ST estimator can include the derivative of the sigmoid or not. \cite{raiko2014techniques} derive the version with the derivative of the sigmoid using another interpretation and differentiate it from the original ST estimator in their work. For simplicity, here we always assume that the ST estimator includes the derivative of the sigmoid, which is often essential for achieving the best performance~\citep{raiko2014techniques}.}. The idea is to backpropagate through the thresholding function as if it were the identity function. The estimator is given below:
\begin{equation} \label{eq:deriv_est_st}
 \begin{split}
  \hat{g}_{(\mathcal{ST})} 
  &= f'(x) \nabla_{\theta} \mu_x(\theta) 
  \end{split}
\end{equation}
The estimator resembles the $\hat{g}_{MF}$ term in our estimator in Eq.~\ref{eq:deriv_est}, since for a stochastic binary network the activation function is the mean function. The difference is that this estimator depends on sampled $x_i$ during backpropagation, while our formulation backpropagates the gradient through the mean-field activations. Despite the heuristic derivation, this simple biased estimator works well in practice, most likely due to its overall similarity to the backpropagation algorithm. However, we show that MuProp significantly outperforms ST estimators on certain tasks, which suggests that unbiased estimators are more reliable.  

\subsection{The 1/2 Estimator}
\cite{gregor2014deep} proposed a biased estimator for the Deep AutoRegressive Network (DARN), which we refer to as the ``1/2'' estimator due to its particular choice of the baseline.
Like the Straight-Through estimator, the 1/2 estimator is also specialized for stochastic binary networks. 
For models with only one stochastic unit, the 1/2 estimator corresponds to using the Taylor expansion around the sample $x$, evaluated at a fixed point $\bar{x}$, as the baseline, such that $h(x) = f(x) + f'(x)(\bar{x}-x)$. Since the expectation over $h(x)$ cannot be computed analytically and $\bar{x}$ cannot be chosen to make the baseline mean $0$ for any arbitrary function $f$, this estimator is biased.  Critically, the 1/2 estimator is derived for models with only one unit but is applied to large models by treating each unit as the sole unit of the model. Furthermore, the 1/2 estimator is not well-justified in the multi-layer case. However, backpropagating the gradient estimator derived in the single-layer case is shown to work well for stochastic binary neurons. The basic form of the estimator is given by
\begin{equation} \label{eq:deriv_est_darn}
 \begin{split}
  \hat{g}_{(1/2)} 
  &=\nabla_\theta \log p_\theta(x)\cdot (f'(x)^T\cdot (x-\bar{x}))
  \end{split}
\end{equation}
If $x$ consists of binary random variables, the expression can be further simplified by using $\bar{x}=1/2$ (see \citep{gregor2014deep} for justification); however, for non-quadratic $f$, the estimator is biased:
\begin{equation} \label{eq:deriv_est_darn_binary}
 \begin{split}
  \hat{g}_{(1/2)} 
  &= \frac{f'(x) \nabla_{\theta} \mu_x(\theta)} { 2p_\theta(x)}
  \end{split}
\end{equation}
If $x$ consists of multinomial random variables with $k$ categories, no sensible $\bar{x}$ has been proposed. Since 1/2 estimator is only tested on binary variables, we experimented with three choices of $\bar{x}$: 1/2, 1/$k$, and $\mu_x(\theta)$. Let $\mu_x(\theta) = \text{softmax}(\theta)$ be the mean of the multinomial units ($\theta \in \mathbb{R}^k$), and $x\in\mathbb{R}^k$ is one-hot encoding of the categorical values. Then the 1/2 estimator for the multinomial case becomes
\begin{equation} \label{eq:deriv_est_darn_multi}
 \begin{split}
  \hat{g}_{(1/2)} 
  &= \frac{(f'(x)^T\cdot(x-\bar{x}))( x^T\cdot \nabla_{\theta}\mu_x(\theta)) } { \mu_x(\theta)}
  \end{split}
\end{equation}

\begin{table}[ht]
 \centering
 \small
 \begin{tabular}{c|c c c }
 Estimator & Unbiased? & Require baseline? & Use Cost Gradient? \\
 \hline
  LR	& Yes	& Yes	& No \\
 ST	& No	& No	& Yes \\
 1/2	& No	& No	& Yes \\
 MuProp & Yes & Yes & Yes \\
   \end{tabular}
\caption{Summary of Gradient Estimators for Stochastic Discrete Variables}
\label{tab:grads}
\end{table}

\section{Experiments}
\label{sec:expe}
We compare the LR, ST, and 1/2 estimators with the MuProp estimator on tasks that use a diverse set of network architectures. For the LR and the MuProp estimators, `-C', `-VN', and `-IDB' indicate constant mean baseline, variance normalization, and input-dependent baseline respectively. The first task does not make use of an inference network and involves direct optimization of an approximation to the expected objective. The second task involves training a sigmoid belief network jointly with an inference network by maximizing the variational lower bound on the intractable log-likelihood. MuProp performs consistently well at both tasks.

\subsection{Structured Output Prediction}
\label{sec:sop}

\begin{figure}
\centering
\begin{subfigure}{.65\textwidth}
  \centering
  \includegraphics[width=.99\linewidth]{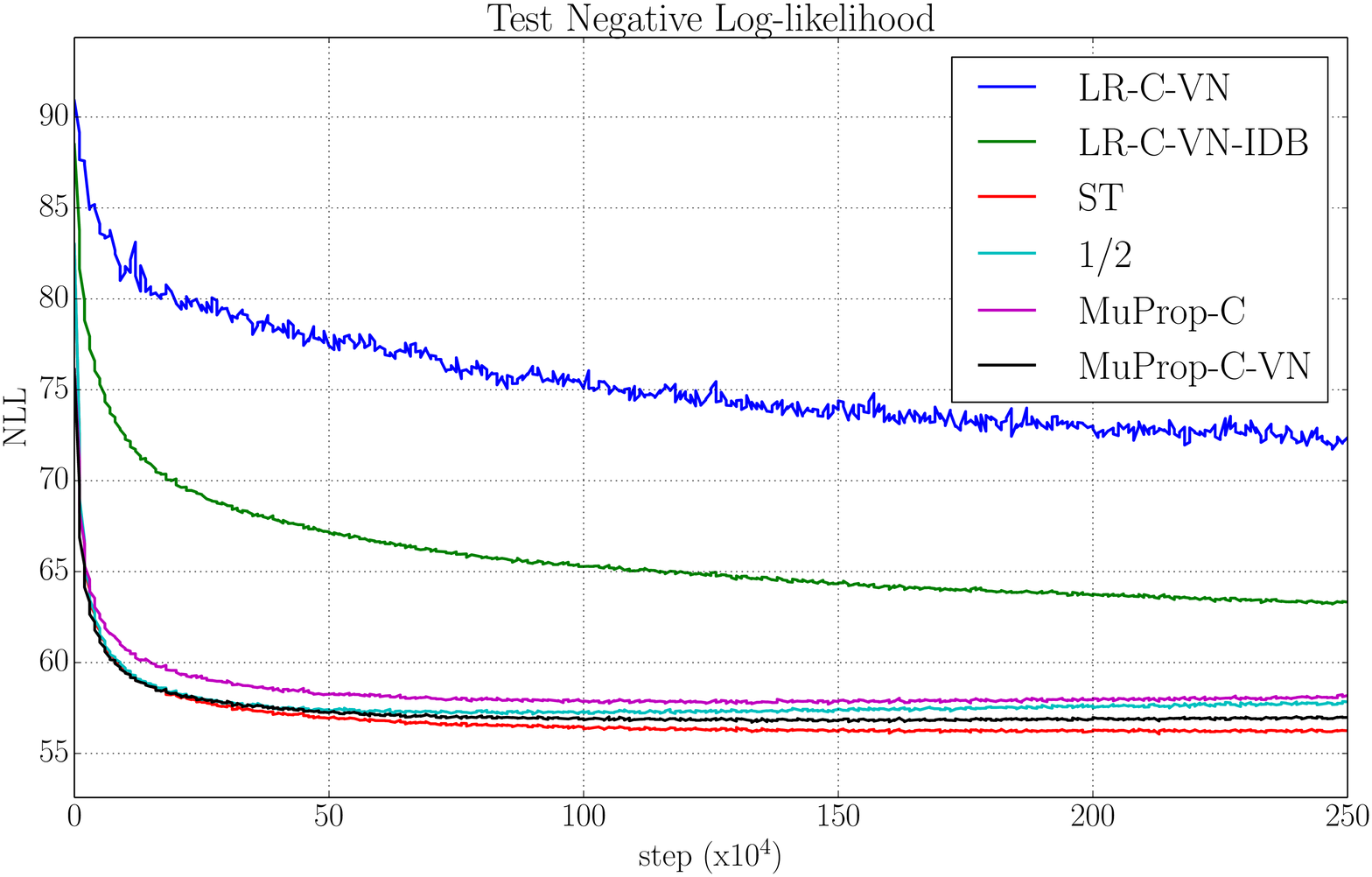}
  \caption{Test negative log-likelihood (NLL)}
  \label{fig:sop}
\end{subfigure}%
\begin{subfigure}{.35\textwidth}
  \centering
  \includegraphics[width=.99\linewidth]{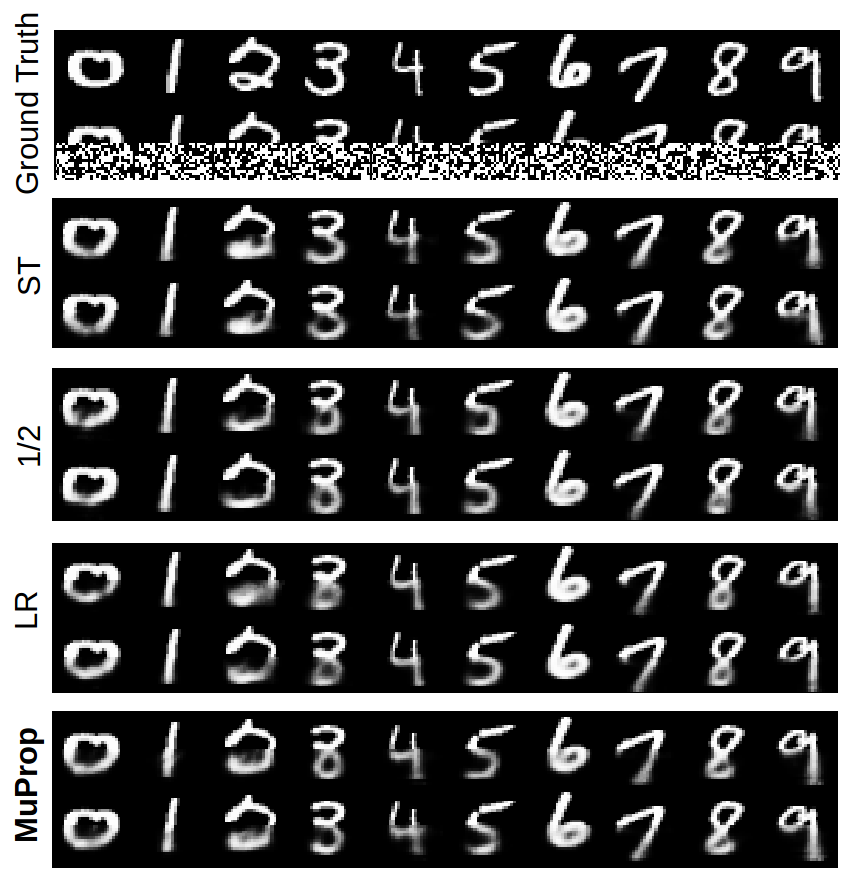}
  \caption{Test samples}
  \label{fig:sop_imp}
\end{subfigure}
\caption{Results for MNIST imputation dataset. MuProp is the only unbiased estimator that can compete with the biased estimators such as ST and 1/2.}
\label{fig:sop_all}
\end{figure}

In this experiment, we follow the setup proposed by \cite{raiko2014techniques}. The two tasks are to predict the lower half of an MNIST digit given the top half, and to predict multiple facial expressions from an average face using Toronto Face dataset (TFD); the output distribution in both tasks exhibits complex multi-modality. For MNIST, the output pixels are binarized using the same protocol as in \citep{raiko2014techniques}. Given an input $x$, an output $y$, and stochastic hidden variables $h$, the objective is to maximize $\mathbb{E}_{h_i\sim p_\theta(h|x)}\left[\log \left(\frac{1}{m} \sum_{i=1}^m p_\theta(y|h_i)\right)\right]$, an importance-sampled estimate of the likelihood objective \citep{raiko2014techniques,burda2015importance}. We use $m=1$
for training, and $m=100$ for validation and testing. For MNIST, a fixed learning rate is chosen from \{$0.003,0.001,..,0.00003$\}, and the best test result is reported for each method. For the TFD dataset, the learning rate is chosen from the same list, but each learning rate is 10 times smaller. We used a momentum of $0.9$ and minibatches of size $100$. The input-dependent baseline (IDB) of \cite{mnih2014neural} uses both the input and output as its input and has a single hidden-layer of 100 Tanh units. 

\begin{table}[ht]
 \centering
 \small
 \begin{tabular}{c|c c c c c c c}
 Estimator & LR-C-VN-IDB & MuProp-C & MuProp-C-VN & ST & $1/2$ \\
 \hline 
 MNIST 392-200-200-392 &  63.2 & 57.7 & 56.7&56.1 &57.2 \\
 MNIST 392-200-200-200-392 & 100.6 & 62.7 & 57.2 & 54.4 & 55.8 \\
 TFD 2034-200-200-2034 & 38.2 & 29.6 & 28.2 & 27.7 & 27.6 \\ 
 \end{tabular}
\caption{Test negative log-likelihood on MNIST 392-200-200-392 model.}
\label{tab:sop}
\end{table}





Figure~\ref{fig:sop} shows the test cost versus the number of parameter update steps of a typical run, while Table~\ref{tab:sop} summarizes the test performance for the different estimators. Convergence of the LR estimators is significantly slower than that of MuProp, ST, and 1/2. In fact, MuProp with simple mean subtraction is consistently better than LR with all of the variance reduction techniques. For MuProp, adding input-dependent baselines does not lead to any improvement over a simple constant mean baseline and variance normalization, and it is interesting to observe that MuProp's sample-dependent-baseline reduces the variance significantly more than the input-dependent baseline that also utilizes the output $y$. MuProp is slightly worse than ST and 1/2, and this performance gap can be explained by the fact that standard MuProp only uses a single-trunk mean-field pass of the network, making the ``mean'' values at higher layers less correlated with the true conditional distribution. Section~\ref{sec:disc} discusses extensions to address this problem, at the cost of more computation or training an auxiliary network.


Our results show that MuProp significantly outperforms LR and closely matches the performance of ST and 1/2 in terms of both convergence speed and final accuracy, and is the only unbiased estimator that can compete with the biased ST and 1/2 estimators on this task.


\subsection{Variational training of generative models}
\label{sec:dvae}

In the second set of experiments, we apply MuProp to variational training of generative models. The auto-encoding variational Bayes framework \citep{kingma2013auto}, along with similar methods, allows powerful generative models to be trained efficiently by replacing slow iterative inference algorithms with fast feedforward approximate inference networks. The inference networks, which map observations to samples from the variational posterior, are trained jointly with the model by maximizing a common objective. 
This objective is a variational lower bound on the marginal log-likelihood, and it is straight-forward to show that it is a stochastic computation graph with particular choice of cost. 

When the variational distributions over the latent variables have a specific form, such as conditional Gaussian, such models can be trained easily by using the reparameterization trick to backpropagate the gradient through the inference network~\citep{kingma2013auto}. However, Gaussian distributions are not appropriate for all types of data or all models. If we opt to use discrete latent variables, we would have to choose between gradient estimators that are biased but have low variance \citep{gregor2014deep} or are unbiased but higher variance \citep{mnih2014neural}. As MuProp is unbiased and has relatively low variance due to its use of backpropagation for gradient estimation, we expect it to be particularly well-suited for training such models.

\begin{figure}
\centering
\begin{subfigure}{.5\textwidth}
  \centering
  \includegraphics[width=.99\linewidth]{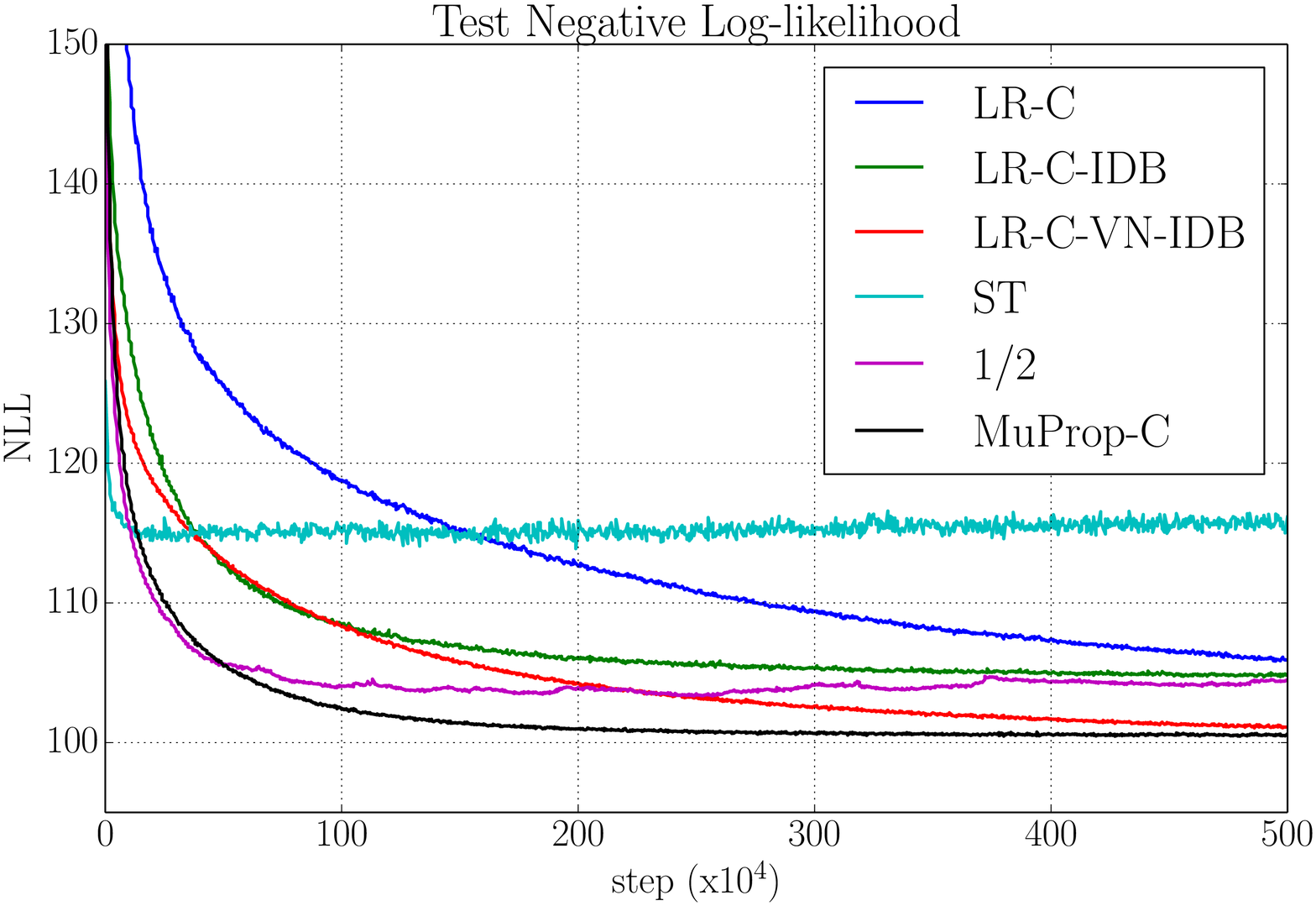}
\end{subfigure}%
\begin{subfigure}{.5\textwidth}
  \centering
  \includegraphics[width=.99\linewidth]{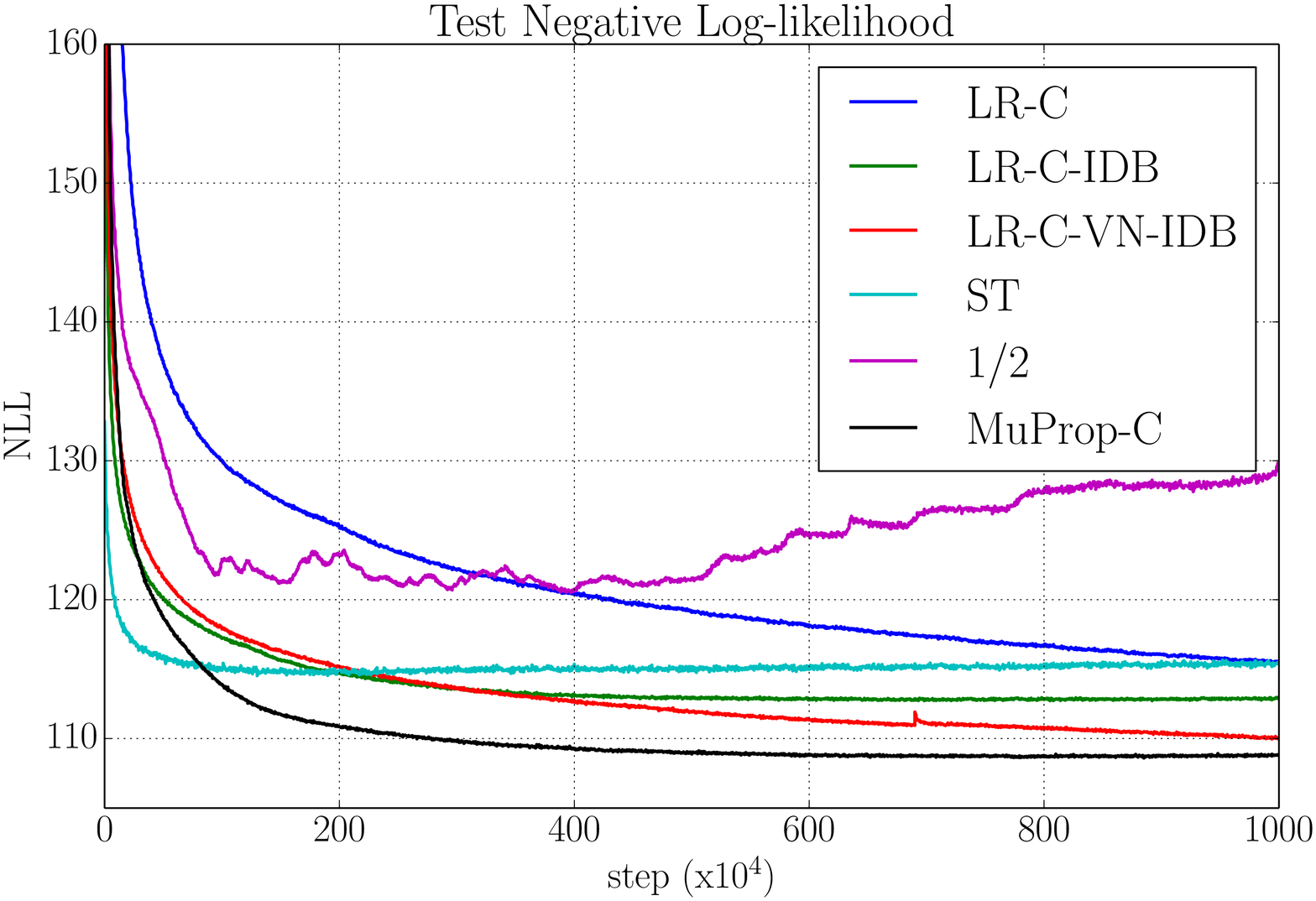}
\end{subfigure}
\caption{Test negative variational lower-bound on 200-200-784 SBN model (left) and 200$\times$10-784 categorical model (right). MuProp outperforms ST and 1/2 and converges faster than LR. }
\label{fig:nvil}
\end{figure}

We will concentrate on training layered belief networks with either Bernoulli or multinomial latent variables. The model in question is equivalent to a sigmoid belief network if it does not have autoregressive connections, or to fDARN if it has autoregressive connections \cite{gregor2014deep}. The multinomial model uses 200 latent variables with 10 categories each ($k=10$), and thus is referred to as the 200$\times$10 model. We applied the models to the binarized MNIST dataset, which consists of 28$\times$28 images of hand-written digits, and is commonly used for evaluating generative models. We trained the models using stochastic gradient descent using momentum of 0.9. The learning rate was selected from \{$0.003,0.001,..,0.000003$\}, and the best test score is reported. For the 1/2 estimator in the categorical model, we found that $\bar{x}$=$1/2$ or $1/k$ worked best, and used the latter value for the results reported.
As we used the same experimental setup and SBN/fDARN models, our results are directly comparable to those obtained using neural variational inference and learning (NVIL) in \citep{mnih2014neural}. Note that LR-C-VN-IDB is our implementation of NVIL.

\begin{table}[ht]
 \centering
 \small
 \begin{tabular}{c|c c c c c }
 Estimator & LR-C-VN-IDB & MuProp-C & MuProp-C-IDB & ST & 1/2 \\
 \hline 
 SBN 200-784 & 113.5 & 113.6& 113.1 & 129.4 & 112.4 \\
 SBN 200-200-784 & 99.7 & 100.4& 100.4 & 119.9 & 103.5 \\
 SBN 200-200-200-784 & 97.5 & 101.1& 98.6 & 116.7 & 99.3 \\
 fDARN 200-784 & 92.1 & 92.9 & 92.7 & 110.2 & 94.2 \\ 
 \hline
 Categorical (200$\times$10)-784 & 107.7 & 108.6 & 107.8 & 114.5 & 120.5 \\
 \end{tabular}
\caption{Negative variational lower-bound on MNIST test data}
\label{tab:vae}
\end{table}

Figure~\ref{fig:nvil} shows sample training curves for typical runs, and Table~\ref{tab:vae} summarizes the results. These results convincingly show that MuProp outperforms all the competing estimators. While the variance-reduced LR estimator (NVIL) performs very well on this task and is on par with MuProp on the final variational lower-bound, MuProp converges significantly faster (e.g. for the categorical model, about 3-4 times faster). This suggests that MuProp (with only mean subtraction) still has significantly less variance than LR on this task, just as in Section~\ref{sec:sop}. For the ST and 1/2 estimators, the story is more interesting. First, their final variational lower-bounds are significantly worse than MuProp or LR. More importantly, they exhibit little consistency in performance. On the SBN models, 1/2 typically outperforms ST, while on the categorical model, the reverse is true. This result, along with the fluctuations observed in the categorical model training curve for 1/2, suggests that such biased estimators can be unreliable for more complex models, and that their performance can vary drastically based on the cost function and model architecture. MuProp, on the other hand, is guaranteed to improve the desired objective because it is unbiased, while having much less variance than LR due to the use of the gradient information from backpropagation.                  
\section{Discussion}
\label{sec:disc}
In this paper, we presented MuProp, which is an unbiased estimator of
derivatives in stochastic computational graphs that combines the statistical efficiency of backpropagation with the correctness of a likelihood ratio method.

MuProp has a number of natural extensions. First, we might consider using other functions for the baseline rather than just the Taylor expansion, which could be learned in a manner that resembles Q-learning~\citep{watkins1992q} and target propagation~\citep{lee2015difference}. In reinforcement learning, fitted Q-functions obtained by estimating the expected return of a given policy $\pi_\theta$ summarize all future costs, and a good Q-function can greatly simplify the temporal credit assignment problem. Combining MuProp with such fitted Q-functions could greatly reduce the variance of the estimator and make it better suited for very deep computational graphs, such as long recurrent neural networks and applications in reinforcement learning.

The second extension is to make $\bar{x}$ depend on samples of its parent nodes, as illustrated by \textit{rollout} procedure in Figure~\ref{fig:scg2}. This could substantially improve performance on deeper networks, where the value from a single-trunk mean-field pass may diverge significantly from any samples drawn with a fully stochastic pass. By drawing $\bar{x}$ using mean-field passes originating at sampled values from preceding layers would prevent such divergence, though at additional computational cost, since the number of mean-field passes would depend on the depth $n$ of the network, for a total of $\mathcal{O}(n^2)$ partial passes through the network. Intuitively, the single mean-field ``chain'' would turn into a ``tree,'' with a sampled trunk and a different mean-field branch at each layer.


\subsubsection*{Acknowledgments}
We sincerely thank Jascha Solh-dickstein, Laurent Dinh, Ben Poole, and Quoc Le for helpful discussions and the Google Brain team for the support. 

{\footnotesize
\bibliography{iclr2016_conference}
\bibliographystyle{iclr2016_conference}
}

\newpage
\section*{Appendix}

\section{Recursive Derivation of MuProp}
Given a model with multiple layers of discrete random variables $x_1...x_n$ and input $x_0$, the general loss function can be written as $L(x_0,\theta)=\mathbb{E}_{p(x_1,...,x_n|x_0,\theta)}[f_\theta(x_{0:n})]$. For simplicity, we show derivation assuming that $p(x_1,...,x_n|x_0,\theta)=\prod_{i=1}^n p(x_{i}|x_{i-1},\theta_{i-1})=\prod_{i=1}^n p_i$ and $f(x_{1:n},\theta)=f(x_n, \theta_n)$, where $\theta=\{\theta_0,...,\theta_n\}$; however, the derivation is easy to extend to any directed acyclic graph. We assume that there is a continuous and differentiable mean function for each $p(x_i|x_{i-1},\theta_{i-1})$, denoted by $\mu_{i}(x_{i-1},\theta_{i-1})=\mathbb{E}_{p(x_{i}|x_{i-1},\theta_{i-1})}[x_i]$. The derivation is based on the recursive first-order Taylor expansion around some arbitrary fixed points $\bar{x}_{1:n}$, which we detail later. We use $\mu'_i$ and $f'$ to denote $\frac{\partial \mu_i}{\partial x_{i-1}}$ and $\frac{\partial f}{\partial x_n}$. 

We first derive the gradient for $\frac{\partial L}{\partial \theta_n}$ and $\frac{\partial L}{\partial \theta_{n-1}}$ and then for $\{\frac{\partial L}{\partial \theta_{i}}\}_{i=0:n}$. The final expression for the estimator resembles the classical backpropagation formulation, and allows efficient computation through forward and backward passes. We intentionally avoid the standard matrix algebra notation to provide clean derivations, as the extension is straight-forward. 
\begin{equation} \label{eq:derivN}
 \begin{split}
  \frac{\partial L}{\partial \theta_n} &= \mathbb{E}_{p_{1:n}}[\nabla_{\theta_n} f(x_n,\theta_n)]
  \end{split}
\end{equation}
\begin{equation} \label{eq:derivN1}
 \begin{split}
  \frac{\partial L}{\partial \theta_{n-1}} 
  &= \mathbb{E}_{p_{1:n-1}}[\nabla_{\theta_{n-1}} \mathbb{E}_{p_n}[f(x_n,\theta_n)]]\\
   &= \mathbb{E}_{p_{1:n-1}}[ \mathbb{E}_{p_n}[\nabla_{\theta_{n-1}}\log p_n \cdot f(x_n,\theta_n)]]\\
   &= \mathbb{E}_{p_{1:n-1}}[\mathbb{E}_{p_n}[\nabla_{\theta_{n-1}}\log p_n\cdot[f(x_n,\theta_n)-f(\bar{x}_n,\theta_n)-f'(\bar{x}_n,\theta_n)\cdot (x_n-\bar{x}_n)]]\\
   &\qquad +\mathbb{E}_{p_n}[\nabla_{\theta_{n-1}}\log p_n\cdot[ f(\bar{x}_n,\theta_n) + f'(\bar{x}_n,\theta_n)\cdot (x_n-\bar{x}_n)]]]\\
   &= \mathbb{E}_{p_{1:n-1}}[\mathbb{E}_{p_n}[\nabla_{\theta_{n-1}}\log p_n\cdot R_n]+f'(\bar{x}_n,\theta_n)\nabla_{\theta_{n-1}}\mathbb{E}_{p_n}[x_N]]\\
   &= \mathbb{E}_{p_{1:n-1}}[\mathbb{E}_{p_n}[\nabla_{\theta_{n-1}}\log p_n\cdot R_n]+f'(\bar{x}_n,\theta_n)\nabla_{\theta_{n-1}}\mu_{n}(x_{n-1},\theta_{n-1})]\\
  \end{split}
\end{equation}

where we define the residuals $R_i=\mu_{i+1}(x_i,\theta_i) - \mu_{i+1}(\bar{x}_i,\theta_i) - \mu'_{i+1}(\bar{x}_i,\theta_i)\cdot (x_i - \bar{x}_i) $ and let $\mu_{n+1}(x_n,\theta_n)=f(x_n,\theta_n)$. Applying the same derivation recursively, we get:

\begin{equation} \label{eq:deriv_i}
 \begin{split}
 \frac{\partial L}{\partial \theta_i} 
 &= \mathbb{E}_{p_{1:n}}[\nabla_{\theta_i}\log p_{i+1} \cdot R_n] + f'(\bar{x}_n,\theta_n)\cdot \mathbb{E}_{p_{1:n-1}}[\nabla_{\theta_i}\log p_{i+1} \cdot \mu_n(x_{n-1},\theta_{n-1}) ]\\ 
 &= \mathbb{E}_{p_{1:n}}[\nabla_{\theta_i}\log p_{i+1} \cdot R_n] + f'(\bar{x}_n,\theta_n)[\mathbb{E}_{p_{1:n-1}}[\nabla_{\theta_i}\log p_{i+1} \cdot R_{n-1} ]\\
 &+ \mu'_n(\bar{x}_{n-1},\theta_{n-1})[ \mathbb{E}_{p_{1:n-2}}[\nabla_{\theta_i}\log p_{i+1} \cdot R_{n-2}]+ ...+\mu'_{i+2}(\bar{x}_{i+1},\theta_{i+1})\cdot \nabla_{\theta_i}\mu_{i+1}(x_{i},\theta_i)]]...] \\
 &= \mathbb{E}_{p_{1:n}}[\nabla_{\theta_i}\log p_{i+1}\cdot (\sum_{k=i+1}^{n}R_k\prod_{j=k+2}^{n+1}\mu'_{j}(\bar{x}_{j-1},\theta_{j-1}))] \\
 &+ \mathbb{E}_{p_{1:i}}[(\prod_{j=i+2}^{n+1}\mu'_{j}(\bar{x}_{j-1},\theta_{j-1}) )\cdot \nabla_{\theta_i} \mu_{i+1}(x_i,\theta_i)]
  \end{split}
\end{equation}
where we slightly abuse the notation and assume $\prod_{j=n+2}^{n+1}(\mu'_j)=1$. The estimator expression is: 
\begin{equation} \label{eq:deriv_est_orig}
 \begin{split}
  \hat{\frac{\partial L}{\partial \theta_i}}_{(\mathcal{\mu})} 
  &= \nabla_{\theta_i}\log p_{i+1}\cdot (\sum_{k=i+1}^{n}R_k\prod_{j=k+2}^{n+1}\mu'_{j}(\bar{x}_{j-1},\theta_{j-1})) + (\prod_{j=i+2}^{n+1}\mu'_{j}(\bar{x}_{j-1},\theta_{j-1}) )\cdot \nabla_{\theta_i} \mu_{i+1}(x_i,\theta_i) 
  \end{split}
\end{equation}
The formulation assumes a set of fixed points $\bar{x}_{1:n}$, whose only requirements are that $\bar{x}_i$ cannot depend on $x_{i}$. If we choose $\bar{x}_{1:n}$ as given by deterministic mean-field forward pass, the gradient estimator simplifies and recovers Eq.~\ref{eq:deriv_est}. 
\section{MuProp with Automatic Differentiation}
\label{sec:autodiff}
We present a simple algorithm to compute MuProp gradient, taking advantage of the automatic differentiation functionalities. 
Algorithm~\ref{alg:muprop} assumes that the automatic differentiation library provides two functionalities: \texttt{Gradients(cost, inputs)} which computes derivatives of \texttt{cost} with respect to \texttt{inputs}, and \texttt{StopGradient(x)} which returns a node with equal value with \texttt{x} but stops gradient when \texttt{Gradients} is called. It assumes that the graph consists of $n$ stochastic nodes $\{x_i\}_{i=1:n}$ and loss function $f$ which can be the sum of multiple loss functions at different parts of the graph. $\texttt{PARENTS}_{x_i}$ denotes the stochastic parental nodes of $x_i$. \texttt{ForwardPass} builds the symbolic graph whose outputs can be differentiated using \texttt{Gradients}. When stochastic=true, each stochastic node samples the value and wraps the value using \texttt{StopGradient} such that no gradient is propagated through the stochastic operation. stochastic=false builds the mean-field network that is fully differentiable. The algorithm also assumes it uses $m=1$ sample, and in general the run time of the algorithm is $1$ deterministic pass + $m$ stochastic passes through the network. In practice, we augment the algorithm with the variance reduction techniques described in Section~\ref{sec:baseline}. Extensions discussed in Section~\ref{sec:disc} are also easy to include.     
\begin{algorithm}
\setstretch{1.35}
 \caption{Compute MuProp Gradient Estimator}
 \label{alg:muprop}
 \begin{algorithmic}
  \REQUIRE Input: $x_0$, Parameters: $\theta$
  \STATE $\bar{x}_{1:n}, f(\bar{x}_{1:n}) \leftarrow \texttt{ForwardPass}(x_0, \text{ stochastic=false})$
  \STATE  $\{f'(\bar{x}_i)\}_{i=1:n} \leftarrow \texttt{Gradients}(f(\bar{x}_{1:n}),\bar{x}_{1:n})$
  \STATE $x_{1:n}, \mu_{1:n}, f(x_{1:n}), \{\log p_\theta(x_i|\texttt{PARENTS}_{x_i})\}_{i=1:n} \leftarrow \texttt{ForwardPass}(x_0, \text{ stochastic=true})$
  \STATE $c_i \leftarrow \log p_\theta(x_i|\texttt{PARENTS}_{x_i})\texttt{StopGradient}(f(x_{1:n})-f(\bar{x}_{1:n})-f'(\bar{x}_i)^T\cdot(x_i-\bar{x}_i) ) + \mu^T\cdot\texttt{StopGradient}(f'(\bar{x}_i))$
  \STATE $c \leftarrow f(x_{1:n}) + \sum_{i=1}^n c_i $
  \STATE $\hat{g}_{(\mu)} \leftarrow \texttt{Gradients}(c, \theta)$
 \end{algorithmic}
\end{algorithm}

\end{document}